\documentclass[nolayout]{article}
\usepackage[english]{babel}
\usepackage{authblk}
\usepackage{graphicx,xcolor,xargs,twoopt,dsfont}
\usepackage{framed}
\usepackage[normalem]{ulem}
\usepackage[active]{srcltx}
\usepackage{amsmath}
\usepackage{amsthm}
\usepackage{amssymb}
\usepackage{amsfonts}
\usepackage{enumerate}
\usepackage[utf8]{inputenc}
\usepackage{fancyhdr}
\usepackage{geometry}
\usepackage{booktabs}
\usepackage{multirow}
\usepackage{xspace} 
\usepackage{url}      
\usepackage{breakcites}
\usepackage{tikz}
\usetikzlibrary{fit,positioning}
\definecolor{violet}{cmyk}{0.79,0.88,0,0}
\definecolor{lavander}{cmyk}{0,0.48,0,0}
\definecolor{burntblue}{cmyk}{0.86,0.30,0.18,0}
\definecolor{burntorange}{cmyk}{0,0.52,1,0}
\definecolor{burntgreen}{cmyk}{0.62,0.44,0.47,0}
\definecolor{colorproof}{RGB}{80,93,113}
\definecolor{lightr}{RGB}{204,0,0}
\definecolor{palegreen}{cmyk}{0.86,0.30,0.96,0}

\newcommand{\eqsp}{\,}

\newcommandx\filtderiv[2][1=]{
\ifthenelse{\equal{#1}{}}
	{\eta_{#2}}
	{\eta_{#2}^\N}
}

\newcommand{\parvec}{\theta}

\newcommand{\tstatletter}{\kernel{T}}

\newcommandx\tstat[2][1=]{
\ifthenelse{\equal{#1}{}}
	{\tstatletter_{#2}}
	{\tau_{#2}^{#1}}
}
\newcommandx\tstathat[2][1=]{
\ifthenelse{\equal{#1}{}}
	{\tstatletter_{#2}}
	{\widehat{\tau}_{#2}^{#1}}
}

\newcommand{\kernel}[1]{\mathbf{#1}}

\newcommandx{\bk}[2][1=]{ 
\ifthenelse{\equal{#1}{}}
{\overleftarrow{\kernel{Q}}_{#2}}
{\overleftarrow{\kernel{Q}}_{#2}^{#1}}
}

\newcommandx{\bkhat}[2][1=]{ 
\ifthenelse{\equal{#1}{}}
{\widehat{\kernel{Q}}_{#2}}
{\widehat{\kernel{Q}}_{#2}^{#1}}
}

\newcommand{\N}{N}

\newcommandx{\K}[1][1=]{
\ifthenelse{\equal{#1}{}}{{\kletter}}{{\widetilde{\N}^{#1}}}}

\newcommand{\kletter}{\widetilde{\N}}

\def\1{\mathds{1}}

\newcommand{\esssup}[2][]
{\ifthenelse{\equal{#1}{}}{\left\| #2 \right\|_\infty}{\left\| #2 \right\|^2_{\infty}}}

\newcommand{\kiss}[3][]
{\ifthenelse{\equal{#1}{}}{r_{#2|#3}}
{\ifthenelse{\equal{#1}{fully}}{r^{\star}_{#2|#3}}
{\ifthenelse{\equal{#1}{smooth}}{\tilde{r}_{#2|#3}}{\mathrm{erreur}}}}}

\newcommand{\chunk}[4][]%
{\ifthenelse{\equal{#1}{}}{\ensuremath{{#2}_{#3:#4}}}{\ensuremath{#2^#1}_{#3:#4}}
}

\newcommand{\kissforward}[3][]
{\ifthenelse{\equal{#1}{}}{p_{#2}}
{\ifthenelse{\equal{#1}{fully}}{p^{\star}_{#2}}
{\ifthenelse{\equal{#1}{smooth}}{\tilde{r}_{#2}}{\mathrm{erreur}}}}}

\newcommandx\post[2][1=]{
\ifthenelse{\equal{#1}{}}
	{\phi_{#2}}
	{\phi_{#2}^\N}
}

\newcommandx\posthat[2][1=]{
\ifthenelse{\equal{#1}{}}
	{\widehat{\phi}_{#2}}
	{\widehat{\phi}_{#2}^\N}
}

\newcommand{\adjfunc}[4][]
{\ifthenelse{\equal{#1}{}}{\ifthenelse{\equal{#4}{}}{\vartheta_{#2|#3}}{\vartheta_{#2|#3}(#4)}}
{\ifthenelse{\equal{#1}{smooth}}{\ifthenelse{\equal{#4}{}}{\tilde{\vartheta}_{#2|#3}}{\tilde{\vartheta}_{#2|#3}(#4)}}
{\ifthenelse{\equal{#1}{fully}}{\ifthenelse{\equal{#4}{}}{\vartheta^\star_{#2|#3}}{\vartheta^\star_{#2|#3}(#4)}}{\mathrm{erreur}}}}}

\newcommand{\XinitIS}[2][]
{\ifthenelse{\equal{#1}{}}{\ensuremath{\rho_{#2}}}{\ensuremath{\check{\rho}_{#2}}}}

\newcommand{\rmd}{\ensuremath{\mathrm{d}}}

\newcommand{\ewght}[2]{\ensuremath{\omega_{#1}^{#2}}}

\newcommand{\epart}[2]{\ensuremath{\xi_{#1}^{#2}}}
\newcommand{\filt}[2][]%
{%
\ifthenelse{\equal{#1}{}}{\ensuremath{\phi_{#2}}}{\ensuremath{\phi_{#1,#2}}}%
}

\newcommand{\sumwght}[2][]{%
\ifthenelse{\equal{#1}{}}{\ensuremath{\Omega_{#2}}}{\ensuremath{\Omega_{#2}^{(#1)}}}}
\newcommand{\sumwghthat}[2][]{%
\ifthenelse{\equal{#1}{}}{\ensuremath{\widehat{\Omega}_{#2}}}{\ensuremath{\widehat{\Omega}_{#2}^{(#1)}}}}

\newcounter{hypH}

\newcommand{\sfr}{\mathsf{r}}
\newcommand{\lag}[1]{\Delta_{#1}}

\newcommand{\smc}{SMC Transformer\xspace}

\newcommand{\mcdrop}{MC Dropout\xspace}

\begin{document}

\title{The Monte Carlo Transformer: a stochastic self-attention model for sequence prediction}
\date{} 

\author[$\dag\top$]{Alice Martin\footnote{This action benefited from the support of the Chair « New Gen RetAIl » led by l’X – \'Ecole Polytechnique and the Fondation de l’\'Ecole Polytechnique, sponsored by CARREFOUR.}}
\author[$\top$]{Charles Ollion}
\author[$\perp$]{Florian Strub}
\author[$\dag$]{Sylvain Le Corff}
\author[$\mp$]{Olivier Pietquin}
\affil[$\dag$]{{\small  Samovar, T\'el\'ecom SudParis, D\'epartement CITI, TIPIC, Institut Polytechnique de Paris, France.}}
\affil[$\top$]{{\small CMAP, UMR 7641, École Polytechnique, CNRS, Institut Polytechnique de Paris, France.}}
\affil[$\mp$]{{\small  Google Research, Brain Team.}}
\affil[$\perp$]{{\small  DeepMind.}}

\lhead{Alice Martin et al.}
\rhead{The Monte Carlo Transformer}

\maketitle

\begin{abstract}
This paper introduces the Sequential Monte Carlo Transformer, an original approach that naturally captures the observations distribution in a transformer architecture. The keys, queries, values and attention vectors of the network are considered as the unobserved stochastic states of its hidden structure. This generative model is such that at each time step the received observation is a random function of its past states in a given attention window. In this general state-space setting, we use Sequential Monte Carlo methods to approximate the posterior distributions of the states given the observations, and to estimate the gradient of the log-likelihood. We hence propose a generative model giving a predictive distribution, instead of a single-point estimate. 
\end{abstract}

\section{Introduction}
\label{sec:introduction}

Many critical applications (e.g. medical diagnosis or autonomous driving) require accurate forecasts while detecting unreliable predictions, that may arise from anomalies, missing information, or unknown situations.
While neural networks excel at predictive tasks, they often solely output a single-point estimate, lacking uncertainty measures to assess their confidence about their predictions. To overcome this limitation, an open research question is the design of neural generative models able to output a predictive distribution instead of single point-estimates. First, such distributions would naturally provide the desired uncertainty measures over the model predictions. Secondly, learning algorithms can build upon such uncertainty measurements to improve their predictive performance such as active learning~\cite{active2000} or exploration in reinforcement learning~\cite{geist2010kalman,Noisynets}.  Thirdly, they may better model incoming sources of variability, such as observation noise, missing information, or model misspecification.

On the one hand, Bayesian statistics offer a mathematically grounded framework to reason about uncertainty, and has long been extended to neural networks~\cite{neal2012bayesian,mackay1992practical}. Among recent methods, Bayesian neural networks (BNNs) estimate a posterior distribution of the target given the input variables by injecting stochasticity in the network parameters~\cite{blundell2015weight,chung2015recurrent} or casting dropout as a variational predictive distribution ~\cite{gal2015dropout}. However, such models tend to be overconfident, leading to poorly calibrated uncertainty estimates~\cite{foong2019pathologies}.
On the other hand, concurrent frequentist approaches have been developed to overcome the computational burden of BNNs, by either computing ensembling networks~\cite{huang2017snapshot,igl2018deep} or directly optimizing uncertainty metrics~\cite{pearce2018high}. Yet, such methods suffer their own pitfalls~\cite{ashukha2020pitfalls}. 
Furthermore, few works focused on measuring uncertainty in sequential prediction problems, adapting the underlying techniques to recurrent neural networks~\cite{fortunato2017bayesian,zhu2017deep}. To the best of our knowledge, none of the stated methods have also been applied to transformer networks~\cite{vaswani2017attention}, even though this architecture reported many successes in complex sequential problems~\cite{li2019enhancing,devlin2018bert}.

To that end, we introduce the Sequential Monte Carlo (SMC) recurrent Transformer, which models uncertainty by introducing stochastic hidden states in the network architecture, as in~\cite{chung2015recurrent}. 
Specifically, we cast the transformer self-attention parameters as unobserved latent states evolving randomly through time.  The model relies on a dynamical system, capturing the uncertainty by replacing deterministic self-attention sequences with latent trajectories. However, the introduction of unobserved stochastic variables in the neural architecture makes the log-likelihood of the observations intractable, requiring approximation techniques in the training algorithm. 

In this paper, we propose to use particle filtering and smoothing methods to draw samples from the distribution of hidden states given observations. Standard implementations of Sequential Monte Carlo methods are based on the auxiliary particle filter~\cite{liu:chen:1998,pitt:shephard:1999}, which is a generalization of~\cite{gordon:salmond:smith:1993,kitagawa:1996} and are theoretically grounded by numerous works in the context of hidden Markov models~\cite{delmoral:2004,cappe:moulines:ryden:2005,delmoral2010backward,dubarry:lecorff:2013,olsson2017efficient}.

Fitting the Transformer approach to general state space modeling provides a new promising and interpretable statistical framework for sequential data and recurrent neural networks.
From a statistical perspective, the \smc provides an efficient way of writing each observation as a mixture of previous data, while the approximated posterior distribution of the unobserved states captures the states dynamics. From a practical perspective, the \smc requires extra-computation at training time, but only needs a single forward pass at evaluation as opposed for example to MC dropout methods~\cite{gal2015dropout}.

We evaluate the \smc model on two synthetic datasets and five real-world time-series forecasting tasks. We show that the \smc manages to capture the known observation models in the synthetic setting, and outperforms all concurrent baselines when measuring classic predictive intervals metrics on the real-world setting.

\section{Background}
\label{sec:background}

\subsection{Sequential Monte Carlo Methods}
\label{subsec:smc}

In real-world machine learning applications, the latent states of parametric models and the data observations tend to be noisy. Generative models have thus been used to replace these deterministic states with unobserved random variables to consider the uncertainty in the estimation procedure. 
However, this leads to an intractable log-likelihood function of the observed data $X_{1:T}$. Indeed, this quantity is obtained by integrating out all latent variables, which cannot be done analytically. Fortunately, a gradient descent algorithm may still be defined using Fisher's identity to estimate the maximum likelihood~\cite{cappe:moulines:ryden:2005}:
\begin{equation}
\label{eq:fisher}
\!\!\nabla_{\theta} \!\log p_{\theta}(X_{1:T}) \!=\! \mathbb{E}_{\theta}[\nabla_{\theta} \!\log p_{\theta}(\zeta_{1:T},X_{1:T})|X_{1:T}]\eqsp,
\end{equation}
where $\theta$ denotes the unknown parameters of the model, $\zeta_{1:T}$ denotes all the unobserved states, $p_\theta$ the joint probability distribution of the observations $X_{1:T}$ and the latent states and $\mathbb{E}_{\theta}$ the expectation under $p_\theta$. 

Sequential Monte Carlo methods, also called particle filtering and smoothing algorithms, aim to approximate the log-likelihood of the generative models by a set of random samples associated with non-negative importance weights. These algorithms combine two steps: (i) a sequential importance sampling step which recursively updates conditional expectations in the form of Eq~\eqref{eq:fisher}, and (ii) an importance resampling step which selects particles according to their importance weights. Following Eq~\eqref{eq:fisher},  $\nabla_{\parvec}  \log p_{\theta}(X_{1:T})$ is then approximated by a weighted sample mean of the form
\begin{equation}
\label{eq:score:MC:background}
S_{\parvec,T}^M = \sum_{m=1}^{M}\omega_{n}^m \nabla_{\theta} \log p_{\theta}(\xi^m_{1:T},X_{1:T}) \eqsp,
\end{equation}
where $(\omega_{n}^m)_{1\leqslant m\leqslant M}$ are nonnegative importance weights such that $\sum_{m=1}^M\omega_{T}^m = 1$ and where $\xi_{1:T}^m$ are trajectories approximately sampled from the  posterior distribution  of $\zeta_{1:T}$ given $X_{1:T}$ parametrized by $\parvec$. Such approximation of the objective function can be plugged into any stochastic gradient algorithm to find a local minima of $\parvec \mapsto  -\log p_{\theta}(X_{1:T})$. 

\subsection{The Transformer model}
\label{subsec:transformer}
Transformers are transduction networks developed as an alternative to recurrent and convolution layers for sequence modeling~\cite{vaswani2017attention}. They rely entirely on (self)-attention mechanisms~\cite{bahdanau2014neural,DBLP:journals/corr/LinFSYXZB17} to model global dependencies regardless of their distance in input or output sequences.

Formally, given the sequence of observations $(X_s)_{s\geqslant 1}$ indexed by $\mathbb{N}^*$, a transduction model aims at predicting an output $X_s$ for a given index $s$ from input data $X_{-s}$. In transformer networks, self-attention modules first associate each input data $X_s$ with a query $q_s$ and a set of key-value $(k_s,v_s)$, where the queries, keys and values are themselves linear projections of the input:
\begin{equation*}
    q(s) =  W^{q} X_s\;, \quad
    \kappa(s) =   W^{\kappa} X_s\;, \quad
    v(s) =   W^{v} X_s\;\,
\end{equation*}
where $W^{q}$, $ W^{\kappa}$ and $ W^{v}$ are unknown weight matrices. A self-attention score then determines how much focus to place on each input in $X_{-s}$ given $X_s$ computed with a dot-product of queries and keys:
\begin{equation*}
\Pi_s=\mathrm{softmax}(Q_{s}K_{s}^T/\sqrt{r})\,,
\end{equation*}
where each line of $Q_s$ (resp. $K_s)$ are matrices whose rows are the values associated with each query (resp. keys) entries. Finally, the self-attention output vector $Z_s$ is the weighted linear combination of  all values:
\begin{equation*}
Z_s = \Pi_s V_{s},
\end{equation*}
where $V_{s}$ is the matrix whose rows are the values associated with each input data. The transformer uses {\em multi-head} self-attention, where the input data is independently processed by $h$ self-attention modules. This leads to $h$ outputs, then concatenated back together to form the final attention output vector. In the rest of the paper, we will consider transformers with a single one-head attention module.

\section{The SMC Transformer}
\label{sec:smc-transformer}

\subsection{Generative model with stochastic self-attention}
\label{subsec:equations}
In this section, we introduce the \smc, a recurrent generative neural network for sequential data based on a stochastic self-attention model.
For all $1\leqslant s\leqslant t$, we define the (key, queries, values) of a self-attention layer of the \smc as follows:
\begin{equation*}
\begin{split}
q(s) =  W^{q} X_s + \Sigma^{1/2}_{q}\varepsilon_q(s)\\
\kappa(s) =   W^{\kappa} X_s+ \Sigma^{1/2}_{\kappa}\varepsilon_\kappa(s)\\
v(s) =   W^{v} X_s + \Sigma^{1/2}_{v}\varepsilon_v(s)
\\
\end{split}
\end{equation*}
where $(\Sigma_{q},\Sigma_{\kappa},\Sigma_{v})$ are unknown semi definite-positive matrices and $(\varepsilon_q,\varepsilon_\kappa,\varepsilon_v)$ are independent standard Gaussian random vectors in $\mathbb{R}^{\sfr}$. Then, we define the matrix $K(t)$ whose columns are the $\kappa(t-s)$, $1\leqslant s \leqslant  \lag{}$, i.e. the past keys up to time $t-\lag{}$. Then, the attention score used at time $t$ is:
\begin{equation*}
\pi(t) = \mathrm{softmax}\,(q(t-1))^TK(t)/\sqrt{r})\eqsp.
\end{equation*}
Finally, the self-attention vector of the input data is computed, for all $1\leqslant  s \leqslant \lag{}$, as,
\begin{equation}
\label{eq:attention:vector}
z(t) = \sum_{s=1}^{\lag{}}\pi_{s}(t)v(t-s)+ \Sigma^{1/2}_{z}\varepsilon_z(t)\eqsp,
\end{equation}
where $\pi_{s}(t)$ denotes the $s$-th component of $\pi(t)$ (i.e. the self attention weight of the observation $t-s$), $\Sigma_{z}$ is an  unknown semi definite-positive matrix and $\varepsilon_z$ are independent standard Gaussian random vectors in $\mathbb{R}^{\sfr}$. Therefore, conditionally on past keys, queries and values, $z(t)$ is a Gaussian random variable with mean  
$\mu(t) = \sum_{s=1}^{\lag{}}\pi_{s}(t)v(t-s)$ and covariance matrix $\Sigma_{z}$. 
Finally, in a regression framework, the observation model is given by:
\begin{equation*}
X_t =  G_{\eta_{obs}}(z_t) + \varepsilon_t\eqsp,
\end{equation*}
where $G_{\eta_{obs}}$ is a feed-forward neural network, and $\varepsilon_t$ is a centered noise such as a centered Gaussian random vector with unknown variance $\Sigma_{\mathrm{obs}}$.

\begin{figure}
\centering
\includegraphics[width=.5\textwidth]{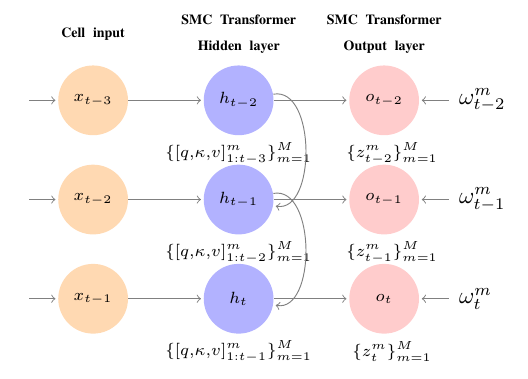}
\caption{Graphical representation of the SMC transformer for sequential data.}
\label{fig:smctransformer}
\end{figure}

By injecting noise in the self-attention parameters of a transformer model, we thus propose a recurrent generative neural network able to predict the conditional distribution of $X_t$ given $\lag{}$ past observations $X_{t-\lag{}:t-1}$ where $1\leqslant \lag{}\leqslant t$. The next section presents the training algorithm to learn the unknown parameters $\theta = (\eta_{obs},\Sigma_{\mathrm{obs}},\{\Sigma,W^{q},W_{}^{\kappa},W_{}^{v}\})$ of this network. A graphical representation of our model is proposed in Figure~\ref{fig:smctransformer}: it describes the dependencies between the latent unobserved states (estimated as a set of $M$ particles), the observations, and the outputs.

\subsection{The training algorithm}
\label{subsec:algo}
In this subsection, we detail how to train the stochastic self-attention model by estimating the objective function, i.e. the negative log-likelihood of the observations, using SMC methods. 
By section~\ref{subsec:equations}, the unobserved state at time $t$ is $\zeta_t = \{z(t),q(t),\kappa(t),v(t)\}$ and the complete-data likelihood may be written:
\begin{equation*}
p_{\theta}(X_{1:T},\zeta_{1:T}) = \prod_{t=1}^{T}p_{\theta}(\zeta_{t}|\zeta_{t-\lag{}:t-1},X_{t-\lag{}:t-1})p_{\theta}(X_{t}|\zeta_{t-\lag{}:t},X_{t-\lag{}:t-1})\eqsp,
\end{equation*}
where by convention if $t-\lag{}\leqslant 1$ then $u_{t-\lag{}:s} = u_{1:s}$. The associated probability density function in the regression setting is: 
\begin{equation*}
p_{\theta}(X_{t}|\zeta_{t-\lag{}:t},X_{t-\lag{}:t-1}) = \varphi_{G_{\eta_{obs}}(z(t)),\Sigma_{\mathrm{obs}}}(X_t)\eqsp,
\end{equation*}
 where $\varphi_{\mu,\Sigma}$ is the Gaussian probability density function with mean $\mu$ and covariance matrix $\Sigma$. By \eqref{eq:fisher} and \eqref{eq:score:MC:background}, the sequential Monte Carlo algorithm approximates $\nabla_{\parvec}  \log p_{\theta}(X_{1:T})$ by a weighted sample mean:
\begin{equation}
\label{eq:score:function}
S_{\parvec,T}^M = \sum_{m=1}^{M}\omega_{T}^m \sum_{t=1}^{T} \big[\nabla_{\parvec} \log p_{\theta}(\xi^m_{t}|\xi^m_{t-\lag{}:t-1},X_{t-\lag{}:t-1}) + \nabla_{\parvec} \log  p_{\theta}(X_{t}|\xi^m_{t-\lag{}:t},X_{t-\lag{}:t-1})\big]\eqsp,
\end{equation}
where the importance weights  $(\omega_{T}^m)_{1\leqslant m\leqslant M}$ and the trajectories $\xi_{1:T}^m$ are sampled according to the particle filter described below. In this paper, we thus estimate the recurrent transformer parameters based on a gradient descent using $S_{\parvec,T}^M$. All parameters related to the noise (i.e., covariance matrices) are estimated using an explicit Expectation Maximization (EM) update~\cite{dempster:laird:rubin:1977} each time a batch of observations is processed as detailed in Appendix A.

\paragraph{Particle filtering/smoothing algorithm.} For all $t\geqslant 1$, once the observation $X_t$ is available, the weighted particle sample $\{(\ewght{t}{m},\epart{1:t}{m})\}_{m=1}^{\N}$ is transformed into a new weighted particle sample. This update step is carried through in two steps, \emph{selection} and \emph{mutation}, using the auxiliary sampler introduced in \cite{pitt:shephard:1999}. New indices and particles $\{ (I_{t+1}^{m}, \epart{t+1}{m}) \}_{m = 1}^\N$ are simulated independently as follows:
\begin{enumerate}
\item Sample $I_{t+1}^{m}$ in $\{1,\ldots,\N\}$ with probabilities proportional to $\{ \ewght{t}{j}\}_{1\leqslant j\leqslant \N}$.
\item Sample $\epart{t+1}{m}$ using the model introduced in Section~\ref{subsec:equations} with the resampled trajectories.
\end{enumerate}
For any  $m \in\{1, \dots, \N\}$, the ancestral line  is updated as follows $\epart{1:t+1}{m} = (\epart{1:t}{I_{t+1}^{m}},\epart{t+1}{m})$ and is associated with the  importance weight defined by
\begin{equation}
\label{eq:weight-update-filtering}
    \ewght{t+1}{m} \propto p_{\theta}(X_{t+1}|\xi^m_{t+1-\lag{}:t+1},X_{t+1-\lag{}:t})\eqsp.
\end{equation}
Therefore, $\omega_{t+1}^m \propto \exp\{-\|X_{t+1}-G_{\eta_{obs}}(z_{t+1}^m)\|^2_{\Sigma_{\mathrm{obs}}}/2\}$ in a regression setting.

This procedure introduced in \cite{kitagawa:1996} (see also \cite{delmoral:2004} for a discussion) approximates the joint smoothing distributions of the latent states given the observations using the genealogy of the particles produced by the auxiliary particle filter.
The genealogical trajectories are defined recursively and updated at each time step with the particles and indices $(\xi^{m}_{k+1},I^{m}_{k+1})$. As a result, at each time step, the algorithm selects an ancestral trajectory by choosing its last state at time $k$, then extended using the newly sampled particle $\xi^{m}_{k+1}$. 

Such algorithm, by maintaining a set of weighted particles and associated genealogical trajectories as an estimation of the \smc's stochastic latent states, allows solving two usual objectives in state-space models: (i) the \textit{state estimation problem}, which aims to recover the latent attention parameter $z_t$ at time $t$ given the observations $X_{1:t}$, and (ii) the \textit{inference problem} which aims at approximating the distribution of $X_t$ given $X_{1:t-1}$. The next section focuses on the latter one, which provides a natural measure of uncertainty for the \smc predictions.

\subsection{Inference and predictive distribution.}

Given the parameter estimate $\widehat{\theta}$ obtained  after the training phase, to solve the inference problem, note that $$
p_{\widehat \theta}(X_t|X_{1:t-1}) = \int p_{\widehat \theta}(X_t,z_{1:t}|X_{1:t-1})\rmd z_{1:t}\eqsp,
$$
which may be approximated using the weighted samples at time $t-1$ by
$$
\widehat{p}^M_{\widehat \theta}(X_t|X_{1:t-1}) =\sum_{m=1}^{M}\omega_{t-1}^m \int p_{\widehat \theta}(X_t|z_t) p_{\widehat \theta}(z_t|\xi^m_{1:t-1},X_{1:t-1})\rmd z_t\eqsp.
$$
A Monte Carlo approximation of the predictive probability $\widehat{p}^M_{\widehat \theta}(X_t|X_{1:t-1})$ may be obtained straightforwardly by sampling from $ p_{\widehat \theta}(z_t|\xi^m_{1:t-1},X_{1:t-1})$. With such estimate, we obtain as a predictive distribution a mixture  $x_t\mapsto \sum_{m=1}^{M}\omega_{t-1}^m \varphi_{G_{\eta_{\mathrm{obs}}}(\hat z_{t}^m),\Sigma_{\mathrm{obs}}}(x_t)$,
where each mixture component $\hat z_{t}^m$ is sampled from $ p_{\widehat \theta}(z_t|\xi^m_{1:t-1},X_{1:t-1})$, i.e. from one of the $M$ particles outputted by the \smc. Obtaining one sample $\hat{x}_t$ from this distribution amounts to (i) sampling a particle index $m$ with probabilities $\omega_{t-1}^m$, $1\leqslant m \leqslant M$, (ii) sampling  $\hat z_{t}^m$  from $ p_{\widehat \theta}(z_t|\xi^m_{1:t-1},X_{1:t-1})$, and (iii) sampling $\hat{x}_t$ from the Gaussian distribution with mean  $G_{\eta_{\mathrm{obs}}}(\hat z_{t}^m)$ and variance $\Sigma_{\mathrm{obs}}$.
This Monte Carlo estimate can be extended straightforwardly to predictions at future time steps.  

Such inference procedure is computationally efficient as it is based directly on the $M$ particles outputted by one single forward pass of the model on the test dataset. It also offers a flexible framework to estimate the predictive distribution. Indeed, the algorithm can be extended to more sophisticated estimation methods than a simple Monte Carlo estimate, that could improve both the predictive performance of the \smc, and its uncertainty estimation. We leave such a possibility for future works.

\section{Experiments}
\label{sec:experiments}

\subsection{Experimental Settings and Implementation details}\label{sec:experimental_setting}
To evaluate the performances of the \smc, we designed two experimental protocols.
First, we create two synthetic datasets with known observation models: the goal is to assess whether the \smc can capture the true distribution of the observations. Second, we evaluate our model on several real-world datasets on time-series forecasting problems while measuring classic predictive intervals metrics.

For each experiment, we compare the \smc with the following baselines: a deterministic LSTM~\cite{hochreiter1997long} and transformer, a LSTM and transformer with \textit{\mcdrop}~\cite{gal2015dropout}, and a Bayesian LSTM~\cite{fortunato2017bayesian}\footnote{We use the implementation from the blitz github library \url{https://github.com/piEsposito/blitz-bayesian-deep-learning}}. 
As mentioned in Section~\ref{subsec:equations}, we implement one-layer transformers with a single self-attention module, and the projection $G_{\eta_{obs}}$ is a point-wise feed-forward network with layer normalization~\cite{ba2016layer} and residual connections as in \cite{ba2016layer}. To ensure full differentiability of the \smc's algorithm, we applied the {\em reparametrization trick}~\cite{Kingma2014AutoEncodingVB} on the Gaussian noises of the self-attention's random variables. Additional details about datasets, models, and training algorithm's hyper-parameters are provided in Appendix B. 

\subsection{Estimating the true variability of the observations on synthetic data}
We design two synthetic auto-regressive time-series with a sequence length of 24 observations. For {\bf model I}, one data sample $X=(X_0,X_1,...,X_{24})$ is drawn as follows, $X_0 \sim \mathcal{N}(0,\,1)$ and for $t\geqslant 0$:
$$
X_{t+1} = \alpha X_t + \sigma \varepsilon_{t+1}\eqsp,
$$
where $(\varepsilon_{t})_{1\leqslant t\leqslant 24}$ are i.i.d standard Gaussian variables independent of $X_0$. For {\bf model II},  the law of a new observation given the past is multimodal and drawn as follows, $X_0 \sim \mathcal{N}(0,\,1)$ and for $t\geqslant 0$:
$$
X_{t+1} = \alpha U_{t+1} X_t + \beta (1-U_{t+1}) X_t + \sigma \varepsilon_{t+1}\eqsp,
$$
where $(\varepsilon_{t})_{1\leqslant t\leqslant 24}$ are i.i.d standard Gaussian variables independent of $X_0$ and $(U_{t})_{1\leqslant t\leqslant 24}$ are i.i.d Bernoulli random variables with parameter $p$ independent of $X_0$ and of variance $(\varepsilon_{t})_{1\leqslant t\leqslant 24}$. In {\bf model I}, the dataset is sampled with $\alpha = 0.8$ and $\sigma^2 = 0.5$. In {\bf model II}, the dataset is sampled with $\alpha = 0.9$, $\beta =0.6\alpha$, $p=0.7$ and $\sigma^2 = 0.3$. 

\begin{table}
\caption{Mean Square Error of the mean predictions (mse) and Mean Square Error of the predictive distribution (dist-mse) on the test set versus the ground truth, for Model I and II.
Values are computed with a 5-fold cross-validation procedure on each dataset. 
Std values are displayed in parenthesis when stds $\geq 0.01.$.
For the LSTM and Transformer models with \mcdrop, $p$ is the dropout rate.
For the Bayesian LSTM, $M$ is the number of Monte Carlo samples to estimate the ELBO loss~\cite{fortunato2017bayesian}. 
For the \smc, $M$ is the number of particles of the SMC algorithm. 
}
\label{table:mses-synthetic}
\scriptsize
\centering
\begin{tabular}{p{0.2\textwidth}|p{0.1\textwidth}p{0.1\textwidth}|p{0.1\textwidth}p{0.1\textwidth}}
\toprule
    \multirow{2}{*}{Model} &
      \multicolumn{2}{c}{\textbf{Model I}} &
      \multicolumn{2}{c}{\textbf{Model II}} \\
      & {mse} & {dist-mse} & {mse} & {dist-mse}  \\
      \midrule
     \textbf{True Model}  & 0.5 & 0.50(0.03) & 0.3 & 0.35(0.07) \\
           \midrule
      \textbf{LSTM} & \textbf{0.50} & N.A  & 0.32 & N.A \\
      \textbf{Transformer} & 0.52 & N.A  & 0.32 & N.A \\ 
      \midrule
      \textbf{LSTM drop.}  &  &   \\
$p = 0.1$    & 0.48 & 0.004  & 0.32 & 0.003 \\
$p = 0.5$    & 0.53 & 0.03  & 0.33 & 0.02 \\
\midrule
\textbf{Transf. drop.}  &  &  \\
$p = 0.1$    & \textbf{0.50} & 0.02  & 0.31 & 0.03 \\
$p = 0.5$    & 0.52(0.01) & 0.05(0.02)  & 0.33(0.02) & 0.05(0.02) \\
\midrule
\textbf{Bayes. LSTM } &  &  \\
$M=10$  &  0.53(0.01) & 0.03& 0.37(0.01) & 0.04 \\
\midrule
\textbf{SMC Transf.}  &  &  \\
$M=10$  &  0.52 & \textbf{0.49} & \textbf{0.30} & \textbf{0.35}
\\
$M=30$  &  0.49 & 0.52 & 0.34 & \textbf{0.35}
\\
\bottomrule
  \end{tabular}
\end{table}

\begin{figure}
\centering
\includegraphics[width=.75\textwidth]{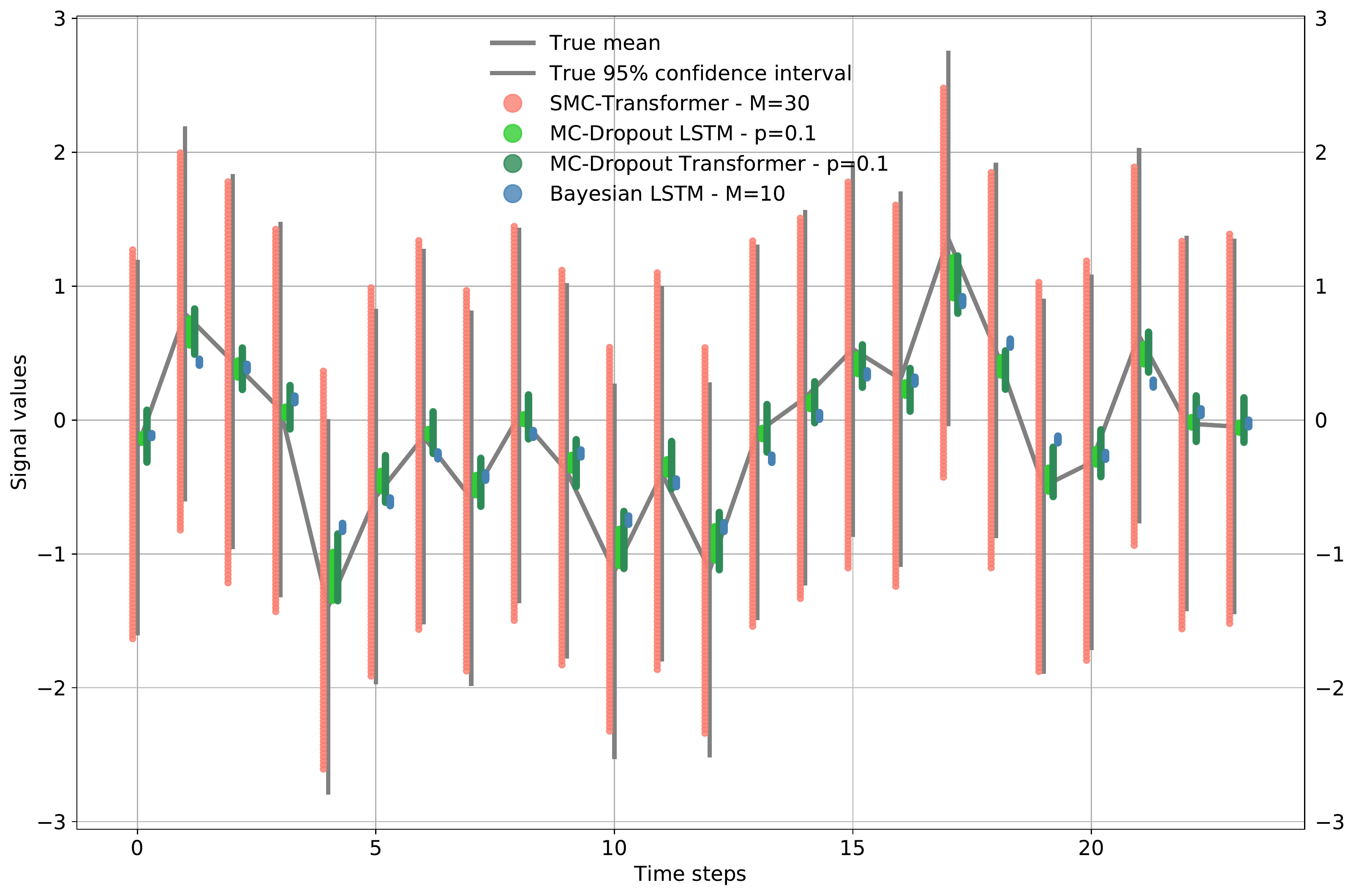}
\caption{Samples distribution on a test example (Model I).}
\label{fig:pred:synthetic}
\end{figure}

Experimental results are summarized in Table~\ref{table:mses-synthetic} and Figure~\ref{fig:pred:synthetic}. 
Table~\ref{table:mses-synthetic} presents two metrics for the two synthetic models. 
First, we compute the Mean Square Error (\textit{mse}) between the true observations and the predicted mean of the observations to measure the model predictive performance. For the \smc, this corresponds to the mean square error between the weighted mean over the particles' predictions and the ground truth. 
Secondly, we refer as \textit{dist-mse}, the empirical estimate of the mean square error of the predictive distribution of $X_{t+1}$ given the $\Delta$ past observations for all time steps $t$. Such an estimate is obtained by generating 1000 samples from the predictive distribution. For the \smc, they are drawn from the SMC estimate of the law of $X_{t+1}$ given $X_{0:t}$. For the baselines, they are drawn by performing 1000 stochastic forward passes on each data sample. For Model I, the \textit{dist-mse} measure is given by $\mathbb{E}[(X_{t+1}-\alpha X_{t})^2|X_t]$, and the true value is $\sigma^2 = 0.5$. For Model II, the measure is $p\mathbb{E}[(X_{t+1}-\alpha X_{t})^2|X_t] + (1-p)\mathbb{E}[(X_{t+1} -\beta X_{t})^2|X_t]$, for which the true value is 0.35. 
In Table~\ref{table:mses-synthetic}, we observe that all models perform similarly when predicting the mean of the observations. Yet, only the \smc manages to capture the true distribution of the observations accurately, with a \textit{dist-mse} measure close to the ground truth. On the other side, both LSTM with \mcdrop and the Bayesian LSTM highly underestimate their variability, as illustrated by the small values of their \textit{dist-mse}.
Such findings are also illustrated in Figure~\ref{fig:pred:synthetic}. We there display the predictive distribution of the different methods versus the true 95\% confidence interval given by the known observation model for the 24 timesteps of a test sample from {\bf Model I}. 
Again, the \smc tends to match the true variability of the observations while concurrent methods clearly underestimate it.

\subsection{Predictive Intervals on real-world time-series.}
 The performance of the stochastic Transformer is evaluated on five real-world sequence predictions problems using the Covid-19\footnote{https://github.com/CSSEGISandData/COVID-19}, Jena weather\footnote{https://www.bgc-jena.mpg.de/wetter/}, the GE stock\footnote{https://www.kaggle.com/szrlee/stock-time-series-20050101-to-20171231} datasets, and the air quality and energy consumption data from the UCI repository\footnote{https://archive.ics.uci.edu/ml/datasets}. Each dataset was split between a train dataset containing 70\% of the data samples, and a validation and test sets containing an equal number of the remaining 15\% of the data samples. Further dataset details are available in Appendix B.

For each of these time-series, we both perform \textit{unistep forecast} and \textit{multistep forecast}. Unistep forecast estimates the conditional distribution $p_{\widehat \theta}(X_t|X_{1:t-1})$ for every timestep of every test sample.
  The multistep forecast estimates the predictive distribution $p_{\widehat \theta}(X_{\tau_H+t}|X_{1:\tau_H})$, with $1\leqslant t \leqslant \tau_F$ on each test sample, given a frozen history of $\tau_H$ timesteps and a number $\tau_F$ future timesteps to predict.

 \paragraph{Predictive intervals metrics.} In real-world time-series, we do not have access to the true distribution of the observations: we thus assess uncertainty by computing the Predicted Interval Coverage Percentage (PICP)~\cite{pearce2018high}.
 For any time step $t$ of the test set, a generative model can provide a lower and upper predicted interval (PI) bound, respectively $\hat{x}_{L_t}$ and $\hat{x}_{U_t}$ by sampling $M$ predictions of the predictive distribution at time $t$.
 The PICP~\cite{pearce2018high} of the true observations is then defined as:
$$
\mathrm{PICP} = \frac{1}{n}\sum_{i=1}^n k_i\quad \mathrm{with} \quad k_i = \begin{cases}
    1& \text{if } \hat{x}_{L_i} \leqslant x_i \leqslant \hat{x}_{U_i}\,,\\
    0             & \text{otherwise}\,,
\end{cases}
$$
where the mean is computed over all time steps considered in the test set. The Mean Predicted Interval Width (MPIW) is: 
$$
\mathrm{MPIW} = \frac{1}{n} \sum_{i=1}^n \left(\hat{x}_{U_i} - \hat{x}_{L_i}\right)\,.
$$
If $\hat{x}_L$ and $\hat{y}_U$ represent the predictive bounds of a $(1-\alpha)$ confidence interval, intuitively, we want the associated $\mathrm{PICP}_\alpha$ to capture $1-\alpha$ proportion of the true observations, while having a corresponding $\mathrm{MPIW}_\alpha$ as small as possible.

\paragraph{Results.} Table~\ref{table:picp:mpiw:multistep} presents the tuple $(\mathrm{PICP}_{0.05}, \mathrm{MPIW}_{0.05})$ associated with a 95\% confidence interval when performing multistep forecasting on the five datasets. The unistep forecasting results are reported in Appendix B. Similarly to Section~\ref{sec:experimental_setting}, we also report the \textit{mse} over the test set between the mean predictions and the true observations. For $(\mathrm{PICP}_{0.05}, \mathrm{MPIW}_{0.05})$, the highest $\mathrm{PICP}_{0.05}$ gives the best measure when it is lower than $0.95$; otherwise, the lowest $\mathrm{MPIW}_{0.05}$ gives the best measure, as proposed in \cite{pearce2018high}. Additional results displaying the performances of the deterministic baselines, and others \mcdrop LSTM and Transformer variants are available in Appendix B. 

Again, while all approaches present similar single-point estimate performances as illustrated with the \textit{mse} values. The \smc outperforms the baselines in terms of $\mathrm{PICP}_{0.05}$ and $\mathrm{MPIW}_{0.05}$ for all datasets, except the energy consumption data, for which our approach is slightly outperformed by the \mcdrop LSTM with dropout rate equal to $0.1$.

\begin{figure}
\centering
    \includegraphics[width=0.8\textwidth]{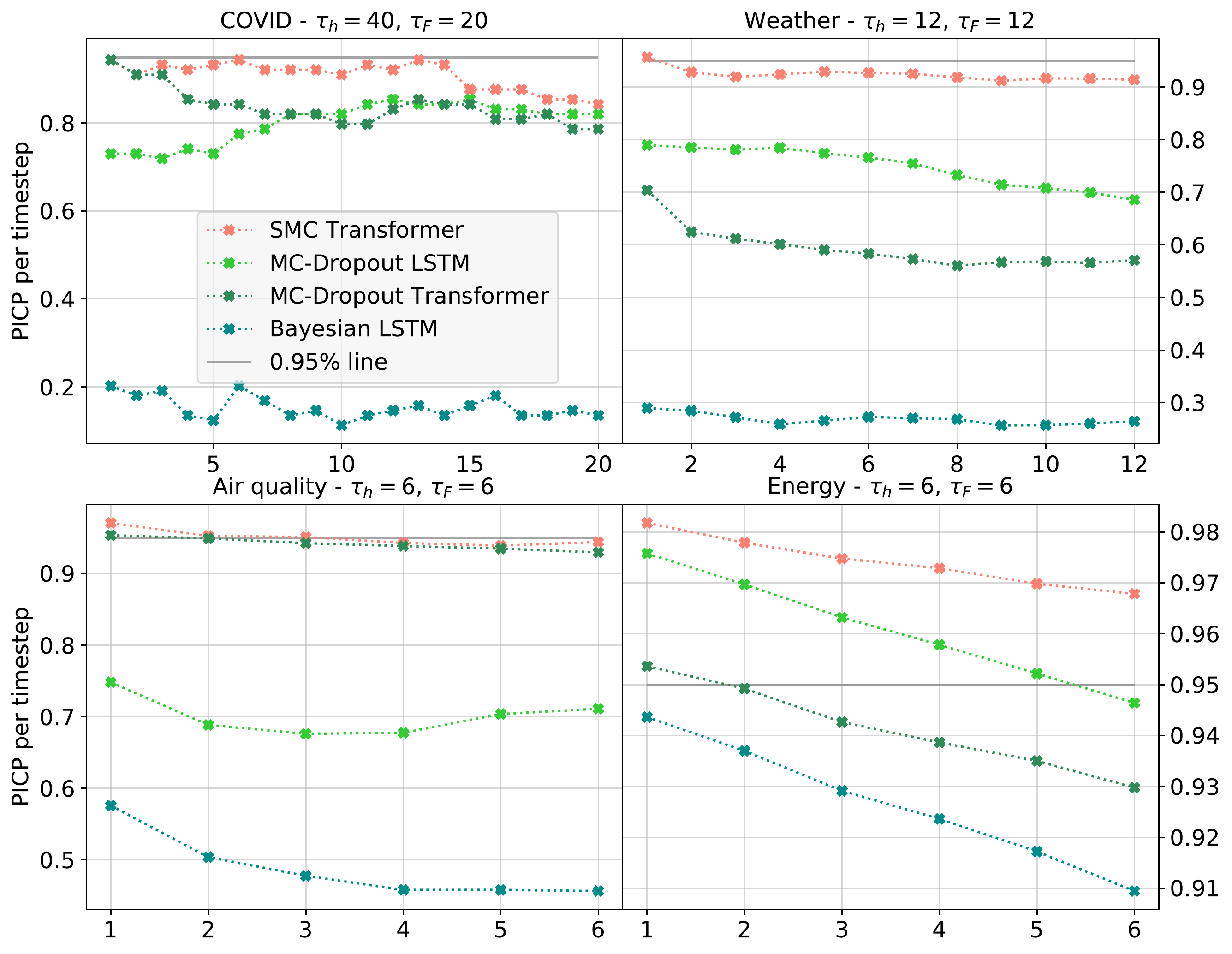}
    \caption{Plot of PICP per timestep when doing multi-step forecast for four of the five datasets. $\tau_H$ represents the number of past timesteps used to predict each of the $\tau_F$ future timesteps.}\label{fig:picp_per_timestep}
    \label{fig:PICP_per_timestep}
\end{figure}

Figure~\ref{fig:picp_per_timestep} represents the evolution of the $\mathrm{PICP}_{0.05}(t)$ per timestep $t$ when doing multistep forecasting for four of the five datasets: the \smc gives higher $\mathrm{PICP}_{0.05}(t)$ and tend to have a stabler PICP evolution over time than the concurrent baselines. Moreover, among the other approaches, there is no evident second best model for predicting uncertainty: sometimes the SMC Transformer is trailed by the \mcdrop Transformer, sometimes by the \mcdrop LSTM. The Bayesian LSTM tends to be particularly overconfident with PICP values often much lower than the ideal 95\% threshold.
As for the \mcdrop models, higher uncertainty measures (obtained generally with higher dropout rates) comes at the expense of predictive performance degradation. 
The discrepancy in uncertainty measures between the \smc and the baselines tends to be higher for datasets with longer sequences, suggesting that our approach is well-suited to model complex structured predictions problems with long-range dependencies. For instance, the stock dataset with a long temporal dependency of 40 past timesteps gives a gap of 11\% between the SMC Transformer and the second best model. However, this gap is only equal to 1\% for the energy dataset, which depends only on 12 past timesteps.

 \begin{table*}
 \centering\hspace*{-5.cm}
\scriptsize
\caption{mse (test loss), PICP and MPIW for multistep forecast. The values in bold correspond to the best performances for each metric.} \label{table:picp:mpiw:multistep}
 \centering\hspace*{-0.6cm}
\resizebox{\textwidth}{!}{
\begin{tabular}{c|c|c|c|c|c|c|c|c|c|c|c} 
\toprule
     & \multicolumn{2}{c|}{{\small Covid}} & \multicolumn{2}{c|}{{\small Air quality}} & \multicolumn{2}{c|}{{\small Weather}} & \multicolumn{2}{c|}{{\small Energy}} & \multicolumn{2}{c}{{\small Stock}} \\
    \hline
  &  mse & picp | mpiw &  mse & picp | mpiw &  mse &  picp | mpiw &  mse & picp | mpiw & mse & picp | mpiw\\
\midrule
\scriptsize
{\small LSTM drop.} &  & & & & & & & \\
$p = 0.1$ & 0.150 & 0.67 | 0.61 & \textbf{0.139} & 0.54 | 0.73 & \textbf{0.089} & 0.62 | 1.15 & 0.07 & \textbf{0.96}\;|\;\textbf{1.12} & \textbf{0.065} & 0.85 | 0.74 \\
$p = 0.5$ & 0.155 & 0.80 | 1.64  & 0.211 & 0.70 | 1.31 & 0.157 & 0.75 | 1.74 & 0.218 & 0.88 | 1.57 & 0.112 & 0.87 | 1.42 \\
\midrule
{\small Transf. drop.} & & & & & & & & \\
$p = 0.1$ & \textbf{0.121} & 0.74 | 0.74  & 0.141 & 0.69 | 1.44 & 0.127 & 0.38 | 0.59 & 0.046 & 0.89 | 0.60 & 0.076 & 0.71 | 0.54 \\
$p = 0.5$ & 0.208 & 0.84 | 1.52  & 0.196 & 0.77 | 1.97 & 0.180 & 0.59 | 1.04 & 0.09 & 0.94 | 1.21 & 0.106 & 0.83 | 0.77 \\
\midrule
{\small Bayesian LSTM} & 0.144 & 0.15 | 0.23 & 0.192 & 0.49 | 0.72 & 0.092 & 0.27 | 0.38 & 0.121 & 0.93 | 1.08 & 0.086 & 0.34 | 0.45 \\
\midrule
{\small SMC Transf.} &  0.128 & \textbf{0.91}\;|\;\textbf{1.85}  & 0.148 & \textbf{0.97}\;|\;\textbf{3.17} & 0.180 & \textbf{0.92}\;|\;\textbf{2.90} & \textbf{0.043} & 0.97 | 1.33 & 0.071 & \textbf{0.98}\;|\;\textbf{1.80} \\
\bottomrule
\end{tabular}
}
\end{table*}

\begin{figure}
\centering
    \includegraphics[width=0.6\textwidth]{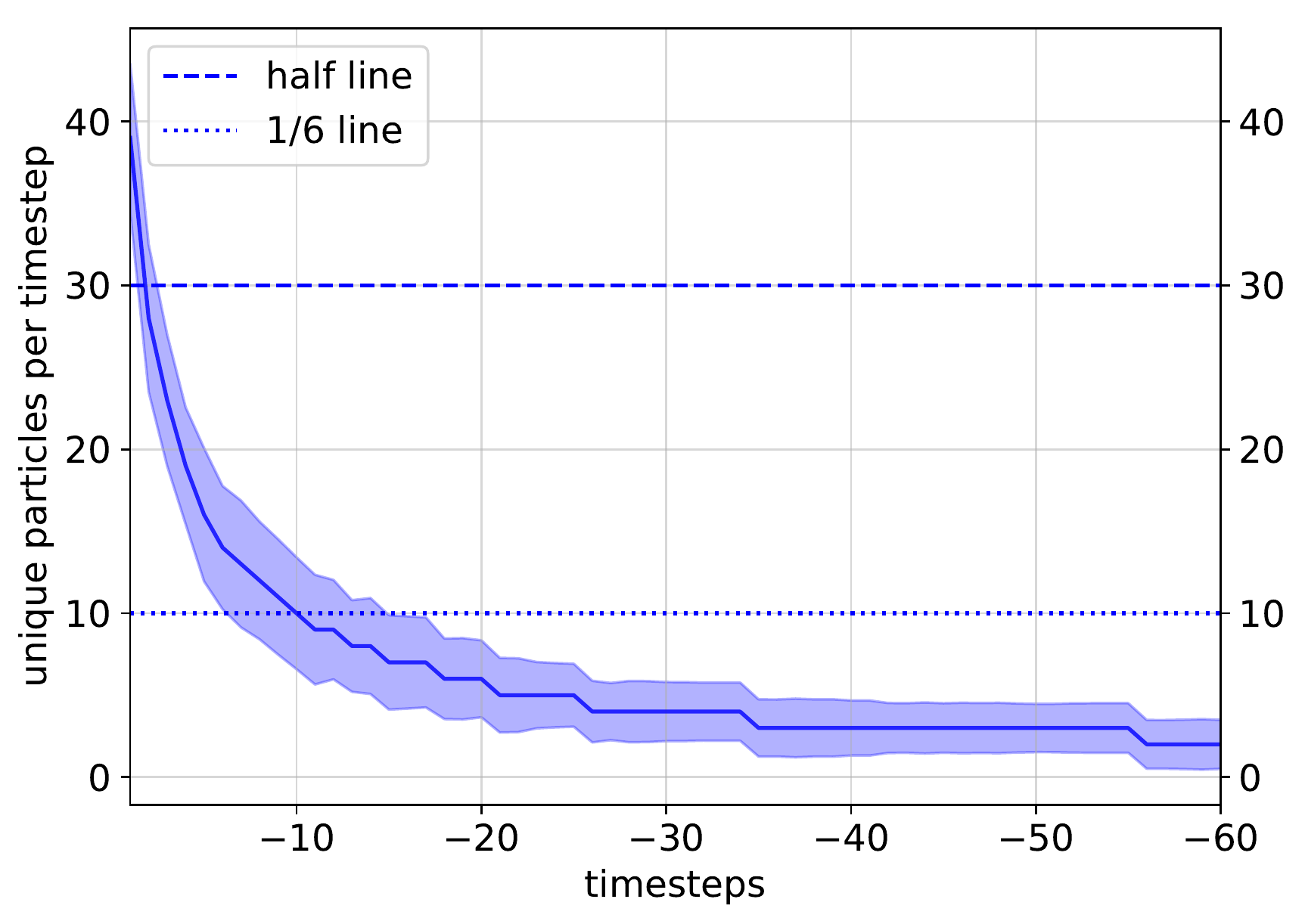}
    \caption{Evolution of the number of unique particles (or non-degenerated particles) over the past timesteps for the resampled trajectories $\xi^m_{t-\lag{}:t-1}$ for a \smc with 60 particles, on the covid dataset. The timesteps are labeled from the most recent one ($t-1$, labeled as -1 on the figure), to the first timestep ($t-\lag{}$, labeled as -60 on the figure).
    The straight line is the average number of particles over the predicted trajectories on the test set, while the shadow area is the corresponding 95\% confidence interval.}
    \label{fig:covid_particles}
\end{figure}

\subsection{Empirical study of the SMC algorithm: particles degeneracy}
\label{subsec:particles:degeneracy}
As highlighted in \cite{kitagawa:1996,kitagawa:sato:2001,fearnhead:wyncoll:tawn:2010,doucet:poyiadjis:singh:2011}, the particle smoother based on the genealogical trajectories suffers from the well known path degeneracy issue. At each time $t\geqslant 1$, the first step to build a new trajectory is to select an ancestral trajectory chosen among $M$ existing trajectories: as the number of resampling steps increases, the number of ancestral trajectories which are likely to be discarded increases.

In Figure~\ref{fig:covid_particles},  we illustrate this degeneracy phenomena on the covid dataset for a \smc with 60 particles. The figure displays the number of unique resampled trajectories $\xi^m_{t-\lag{}:t-1}$ remaining for the 60 timesteps of the sequential process, averaged over the test set.
The further we go in the past, the more the trajectories degenerate: for the first five timesteps, the trajectories are derived from only $2$ unique particles, and only the last ten timesteps present a set of unique particles whom size is superior to one sixth of the original size ($60$).

They are many solutions to improve the approximation $S_{\parvec,T}^M$ and to avoid such phenomena. The easiest to implement is to use the fixed-lag smoother of \cite{olsson:cappe:douc:moulines:2008}: for each $1\leqslant t \leqslant n$, the trajectories $\xi^m_{t-\lag{}:t-1}$ are only resampled up to a few time steps $\tau$ after $t$.
Figure ~\ref{fig:covid_particles} indicates  which value of $\tau$ should be used if we want to keep a sufficiently large number of unique past trajectories. Other approaches based on the decomposition of the smoothing distributions using backward kernels have been widely studied in the hidden Markov models literature \cite{doucet:godsill:andrieu:2000,godsill:doucet:west:2004, delmoral2010backward,olsson2017efficient}. Extending such approaches to deep learning architectures at a reasonable computational cost is a practical challenge. We leave such improvement of the SMC algorithm in our generative model for future works.

\section{Related work}
\label{sec:litterature}
Uncertainty estimation in Deep Learning has sparked a lot of interest from the research community over the last decade, leading to a rich literature on the subject. Such works are usually divided between frequentist approaches~\cite{lakshminarayanan2017simple,huang2017snapshot,pearce2018high,tagasovska2019single,wang2020deeppipe,osband2016deep} and Bayesian ones. Among the latter, Bayesian Neural Networks (BNNs) as defined in \cite{gal2016uncertainty} put a prior distribution over the networks weights~\cite{blundell2015weight,gal2015dropout,khan2018fast,kendall2015bayesian,teye2018bayesian,hernandez2015probabilistic}. However, they suffer from several limitations, one being their inability to correctly assess the posterior distribution~\cite{foong2019pathologies} as illustrated in Section~\ref{sec:experiments}, the other being their computational overhead. Although \mcdrop is one of the most scalable Bayesian inference algorithms, its sampling procedure at inference, that relies on one stochastic forward pass per sample from the predictive distribution, is more computationally expensive than the \smc, for which sampling comes from a Gaussian mixture model directly derived from the particles predictions.  Another Bayesian approach uses neural networks with stochastic hidden states as described below.

\paragraph{RNNs with stochastic latent states.} The \smc is part of an emerging line of research bridging state-space models (historically restricted to simpler statistical models such as Hidden Markov Models or Gaussian Linear Models) and Deep Neural Networks. A few works have proposed recurrent neural networks with stochastic latent states, such as VRAE~\cite{fabius2014variational}, VRNN \cite{chung2015recurrent}, SRNN~\cite{fraccaro2016sequential}, or VHRNN~\cite{deng2020variational}. They use training algorithms that rely on variational inference methods~\cite{blei2017variational} to approximate the intractable posterior distribution over the latent states. Such learning procedures are popular and are computationally efficient, but output a predictive distribution known to be ill-fitted for estimating certain families of distributions, such as for instance multimodal distributions. 
\paragraph{SMC methods and RNNs.} Several SMC algorithms have then been developed to get a better estimator of the marginal likelihood of the observations for such stochastic RNNs, again in a variational inference framework.
\cite{le2017auto,naesseth2017variational,maddison2017filtering} proposed particle filtering algorithms, while \cite{lawson2018twisted,moretti2019particle} developed particle smoothing ones. 
Our work differs from the models and algorithms mentioned above in several ways. First, the training algorithm of the \smc relies on the Fisher's Identity to estimate the gradient of the likelihood of the observations, instead of a variational objective. Secondly, this is the only work proposing: (i) a novel recurrent generative model based on a stochastic self-attention model, and (ii) a novel SMC algorithm to estimate the posterior distribution of the unobserved states, with resampling weights directly depending on the output of the \smc. Finally, while the above works only leverage the SMC algorithm to get a better and lower-variance estimator of the marginal log-likelihood, our work and evaluation protocol focus on uncertainty measurements for sequence prediction problems.

\section{Conclusion}
\label{sec:conclusion}
In this paper, we proposed the \smc, a novel recurrent network that naturally captures the distribution of the observations. This model maintains a distribution of self-attention parameters as latent states, estimated by a set of particles. It thus outputs a distribution of predictions instead of a single-point estimate. Our inference method gives a flexible framework to quantify the variability of the observations.  To our knowledge, this is the first method dedicated to estimating uncertainty in the transformer model, and one of the few focusing on uncertainty quantification in the context of sequence prediction. 
 Moreover, this \smc layer could be used as a "plug-and-play" layer for uncertainty quantification in a deeper neural network encoding sequential data. One limitation of our model is its computational overhead at training time; yet, it can be eased using in particular variants of the SMC algorithm mentioned in section~\ref{subsec:particles:degeneracy}.

\appendix

\section{Details on the training algorithm}

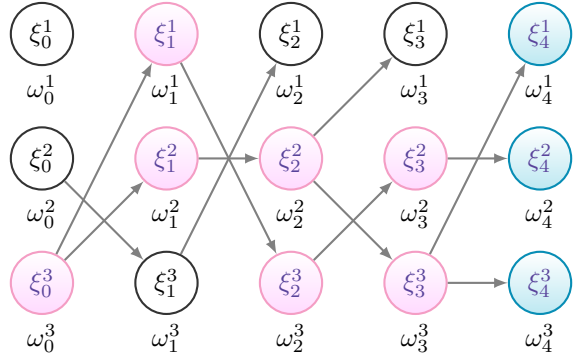
\begin{figure}[h!]
\centering
\begin{tikzpicture}
\tikzstyle{main}=[circle, minimum size = 8mm, thick, draw =black!80, node distance = 8mm]
\tikzstyle{connect}=[-latex, thick, gray]
\tikzstyle{box}=[rectangle, draw=black!100]
\tikzstyle{V3}=[draw =black!80,circle,minimum size = 8mm, thick,node distance = 8mm,lavander,bottom color=lavander!30,top color= white, text=violet]
\tikzstyle{V5}=[draw =black!80,circle,minimum size = 8mm, thick,node distance = 8mm,burntblue,bottom color=burntblue!30,top color= white, text=violet]
\tikzstyle{V4}=[draw =black!80,circle,minimum size = 8mm, thick,node distance = 8mm,burntorange,bottom color=burntorange!30,top color= white, text=violet]
  \node[main] (theta) [label=below:$\omega^2_{0}$] {$\xi_{0}^2$ };
  \node[V3] (theta2) [below=of theta,label=below:$\omega^3_{0}$] {$\xi_{0}^3$ };
  \node[main] (theta3) [above=of theta,label=below:$\omega^1_{0}$] {$\xi_{0}^1$ };
  \node[V3] (z) [right=of theta,label=below:$\omega_{1}^2$] {$\xi_{1}^2$};
  \node[V3] (z2) [above=of z,label=below:$\omega^1_{1}$] {$\xi_{1}^1$};
  \node[main] (z3) [below=of z,label=below:$\omega^3_{1}$] {$\xi_{1}^3$};
  \node[V3] (w) [right=of z,label=below:$\omega^2_{2}$] {$\xi^2_{2}$};
  \node[main] (w2) [above=of w,label=below:$\omega^1_{2}$] {$\xi^1_{2}$};
  \node[V3] (w3) [below=of w,label=below:$\omega^3_{2}$] {$\xi^3_{2}$};
  \node[V3] (x) [right=of w,label=below:$\omega^2_{3}$] {$\xi^2_{3}$};
  \node[main] (x2) [above=of x,label=below:$\omega^1_{3}$] {$\xi^1_{3}$};
  \node[V3] (x3) [below=of x,label=below:$\omega^3_{3}$] {$\xi^3_{3}$};
   \node[V5] (y) [right=of x,label=below:$\omega^2_{4}$] {$\xi^2_{4}$};
  \node[V5] (y2) [above=of y,label=below:$\omega^1_{4}$] {$\xi^1_{4}$};
  \node[V5] (y3) [below=of y,label=below:$\omega^3_{4}$] {$\xi^3_{4}$};
             \path  (w) edge [connect] (x3)
                       (w) edge [connect] (x2)
                     (w3) edge [connect] (x)
                       (z2) edge [connect] (w3)
                       (z3) edge [connect] (w2)
                     (z) edge [connect] (w)
                     (theta2) edge [connect] (z)
                       (theta2) edge [connect] (z2)
                     (theta) edge [connect] (z3)
                     (x) edge [connect] (y)
                       (x3) edge [connect] (y2)
                     (x3) edge [connect] (y3);
\end{tikzpicture}
\caption{Particle filter: $\N=3$, $n=4$.}
\label{fig:pf}
\end{figure}

\paragraph{Particle filtering/smoothing algorithm.} For all $t\geqslant 1$, once the observation $X_t$ is available, the weighted particle sample $\{(\ewght{t}{m},\epart{1:t}{m})\}_{m=1}^{\N}$ is transformed into a new weighted particle sample. This update step is carried through in two steps, \emph{selection} and \emph{mutation},  using the following sampler, see for instance \cite{pitt:shephard:1999}. New indices and particles $\{ (I_{t+1}^{m}, \epart{t+1}{m}) \}_{m = 1}^\N$ are simulated independently as follows:
\begin{enumerate}
\item Sample $I_{t+1}^{m}$ in $\{1,\ldots,\N\}$ with probabilities proportional to $\{ \ewght{t}{j}\}_{1\leqslant j\leqslant \N}$.
\item Sample $\epart{t+1}{m}$ using the model with the resampled trajectories.
\end{enumerate}
In the regression framework of the paper, for any  $m \in\{1, \dots, \N\}$, the ancestral line $\epart{1:t+1}{\ell}$ is updated as follows $\epart{1:t+1}{m} = (\epart{1:t}{I_{t+1}^{m}},\epart{t+1}{m})$ and is associated with the  importance weight defined by
\begin{equation*}
    \ewght{t+1}{m} \propto p_{\theta}(X_{t+1}|\xi^m_{t+1-\lag{}:t+1},X_{t+1-\lag{}:t}) =  \exp\{-\|X_{t+1}-G_{\eta_{obs}}(z_{t+1}^m)\|^2_{\Sigma_{\mathrm{obs}}}/2\} \eqsp,
\end{equation*}
where $\|X_{t+1}-G_{\eta_{obs}}(z_{t+1}^m)\|^2_{\Sigma_{\mathrm{obs}}} = (X_{t+1}-G_{\eta_{obs}}(z_{t+1}^m))^\top\Sigma_{\mathrm{obs}}^{-1} (X_{t+1}-G_{\eta_{obs}}(z_{t+1}^m))$. The algorithm is illustrated in Figure~\ref{fig:pf}: particles at the last time step are in blue and pink particles are the ones which appear in the genealogy of at least one blue particle. White particle have not been selected to give birth to a path up to the last time. In Figure~\ref{fig:pf}, the $N=3$ genealogical trajectories are $\xi^1_{0:4} = (\xi_0^3,\xi_1^2,\xi_2^2,\xi_3^3,\xi_4^1)$, $\xi^2_{0:4} = (\xi_0^3,\xi_1^1,\xi_2^3,\xi_3^2,\xi_4^2)$, $\xi^3_{0:4} = (\xi_0^3,\xi_1^2,\xi_2^2,\xi_3^3,\xi_4^3)$.

\paragraph{The training algorithm}
 The sequential Monte Carlo algorithm approximates $\nabla_{\parvec}  \log p_{\theta}(X_{1:T})$ by a weighted sample mean: 
\begin{equation*}
S_{\parvec,T}^M = \sum_{m=1}^{M}\omega_{T}^m \sum_{t=1}^{T} \left[\nabla_{\parvec} \log p_{\theta}(\xi^m_{t}|\xi^m_{t-\lag{}:t-1},X_{t-\lag{}:t-1}) + \nabla_{\parvec} \log  p_{\theta}(X_{t}|\xi^m_{t-\lag{}:t},X_{t-\lag{}:t-1})\right]\eqsp,
\end{equation*}
where the importance weights  $(\omega_{T}^m)_{1\leqslant m\leqslant M}$ and the trajectories $\xi_{1:T}^m$ are sampled according to the particle filter described below. Thanks to Fisher's identity, this approximation only requires to compute the gradient of state model $\theta \mapsto \log p_{\theta}(\xi^m_{t}|\xi^m_{t-\lag{}:t-1},X_{t-\lag{}:t-1})$ and the gradient of the observation model $\theta\mapsto  \log  p_{\theta}(X_{t}|\xi^m_{t-\lag{}:t},X_{t-\lag{}:t-1})$. There is no need to compute the gradient of the weights $\omega_{T}^m$ which depend on the parameter $\theta$. The loss function used to train the model is therefore
$$
\theta\mapsto  - \sum_{m=1}^{M}\omega_{T}^m \sum_{t=1}^{T} \left[\log p_{\theta}(\xi^m_{t}|\xi^m_{t-\lag{}:t-1},X_{t-\lag{}:t-1}) +  \log  p_{\theta}(X_{t}|\xi^m_{t-\lag{}:t},X_{t-\lag{}:t-1})\right]\eqsp.
$$

In this paper, we propose to estimate all the parameters of the recurrent architecture based on a gradient descent using $S_{\parvec,T}^M$. All parameters related to the noise (the covariance matrices) are estimated using an explicit Expectation Maximization (EM) update \cite{dempster:laird:rubin:1977} each time a batch of observations is processed, see the supplementary materials for all details. For each sequence of observations, the EM update relies on the approximation of the intermediate quantity
$$
\mathbb{E}[\log p_{\theta}(X_{1:T},\zeta_{1:T})|X_{1:T}]= \sum_{t=1}^{T}\mathbb{E}[ \log p_{\theta}(\zeta_{t}|\zeta_{t-\lag{}:t-1},X_{t-\lag{}:t-1}) + \log p_{\theta}(X_{t}|\zeta_{t-\lag{}:t},X_{t-\lag{}:t-1})|X_{1:T}]
$$
by the following particle-based estimator:
$$
Q_{\parvec,T}^M = \sum_{m=1}^{M}\omega_{T}^m \sum_{t=1}^{T}\left[\log p_{\theta}(\xi^m_{t}|\xi^m_{t-\lag{}:t-1},X_{t-\lag{}:t-1}) + \log p_{\theta}(X_{t}|\xi^m_{t-\lag{}:t},X_{t-\lag{}:t-1})\right]\eqsp.
$$
Then, $Q_{\parvec,T}^M $ may be maximized with respect to all covariances to obtain the new estimates. This is a straightforward update which yields for instance for $\Sigma_{\mathrm{obs}}$ for the $p$-th update:
$$  
\Sigma^p_{\mathrm{obs}} = \frac{1}{T}\sum_{m=1}^{M}\omega_{T}^m \sum_{t=1}^{T}(X_{t}-G_{\eta_{obs}}(r_{t}^m))^\top(X_{t}-G_{\eta_{obs}}(r_{t}^m))\eqsp,
$$
where $r_{t}^m$ are the resampled particles at time $t$. The new estimate of $\Sigma_{\mathrm{obs}}$ is then
$$
\widehat{\Sigma}_{\mathrm{obs}} = (1-\eta_p)\widehat{\Sigma}_{\mathrm{obs}} + \eta_p\Sigma^p_{\mathrm{obs}}\eqsp
$$
where $\eta_p$ is a learning rate chosen by the user ($\eta_p = p^{-0.6}$ in the experiments).

\section{Experiments}
\label{sec:experiments:sup}
\subsection{Baselines}
The LSTM and Transformer with \mcdrop\cite{gal2015dropout} have respectively one and two dropout layers. For the LSTM, this layer is just before the output layer. For the Transformer, the dropout layers are inserted in the network architecture as in \cite{vaswani2017attention}. The first dropout layer is after the attention module that computes the output attention vector, and the second one is just before the last layer-normalization layer of the feed-forward neural network that transforms the attention vector. The same dropout rate is kept during training and inference. 

The implementation of the Bayesian LSTM relies on the blitz\footnote{https://github.com/piEsposito/blitz-bayesian-deep-learning} library, following the model from \cite{fortunato2017bayesian}. On the synthetic setting, a hyper-parameter search was performed on $\mathrm{prior_{\sigma_1}}$, the standard deviation of the first component of the gaussian mixture model that models the prior distribution over the network's weights, and on $\mathrm{prior}_{\pi}$, the factor to scale this mixture model, as follows:
$$\mathrm{prior_{\sigma_1}} \in \{0.1, 0.135, 0.37, 0.75, 1, 1.5\}, \quad \mathrm{prior}_{\pi} \in \{0.25, 0.5, 0.75, 1\}\,.$$ The best results in terms of predictive performance were obtained for $\mathrm{prior_{\sigma_1}}=0.1$ and $\mathrm{prior_{\sigma_1}}=1$: we kept these values for the experiments on the real-world datasets. The other hyper-parameters of the Bayesian LSTM were kept at the default values provided by the blitz library. 

\subsection{Hyper-parameters}
\paragraph{Models dimensions.} The LSTM models (deterministic LSTM, \mcdrop  LSTM, Bayesian LSTM) have a number of units in the recurrent layer equal to $32$. The Transformer models (deterministic Transformer, \mcdrop  Transformer, and \smc) have a depth (dimension of the attention parameters) equal to $32$, and a number of units in the feed-forward neural network that transforms the attention vector also equal to $32$. 
\paragraph{Training hyper-parameters.}
For training the \smc and the baselines, we use the ADAM algorithm with a learning rate of 0.001 for the LSTM networks and the original custom
schedule found in [Vaswani et al., 2017] for the Transformer networks. Models were trained for 50 epochs, except the Bayesian LSTM that was trained for 150 epochs. For the two synthetic models, a batch size of $32$ was used. On the real-world setting, batch sizes of 32, 64, 256, 128 and 64 were respectively used for the covid, air quality, weather, energy and stock datasets. 

\subsection{Datasets}
\paragraph{Synthetic data.} The synthetic datasets were generated with 1000 samples: we used 800 of them for training and 100 of them for test and validation. A 5-fold cross-validation procedure was performed at training, to estimate the variability in performance that can be attributed to the training algorithm (by opposition with the measurement of the observations variability, computed with the \textit{dist-mse} metric described in Section 4.2 of the main paper). 

\paragraph{Real-world time series.} We use a 0.7 / 0.15 / 0.15 split for training, validation and test for the real-world datasets.

The covid dataset\footnote{https://github.com/CSSEGISandData/COVID-19} is a univariate time series gathering the daily deaths from the covid-19 disease in 3261 US cities. Cities with less than 100 deaths over the
time period considered were discarded from the dataset, leading to 886 samples in the final dataset, with a sequence length equal to $60$, corresponding to 2 months of observations.

The air quality dataset\footnote{https://archive.ics.uci.edu/ml/datasets/air+quality} gathers hourly responses of a gas multisensor device deployed in an Italian city. It is a multivariate time series with 9 input features: we kept 5 features as target features to be predicted, corresponding to the concentration of 5 chemical gases in the atmosphere. The final dataset has $9,348$ samples, and a sequence length equal to $12$, corresponding to a half day of observations.

The weather dataset\footnote{https://www.bgc-jena.mpg.de/wetter/Weatherstation.pdf} gathers meteorological data from a German weather station. It is a multivariate time series with 4 input and target features (temperature, air pressure, relative humidity, and air density). The final dataset have $420,551$ samples, and a sequence length equal to $24$, corresponding to one day of observations.

The energy dataset\footnote{https://archive.ics.uci.edu/ml/datasets/Appliances+energy+prediction} gathers 10-min measurements of household appliances energy consumption, coupled with local meteorological data. It is a multivariate time series with 28 input features: we kept 20 target features to be predicted. The final dataset have $19,735$ samples, and a sequence length equal to $12$, corresponding to 2 hours of observations.

The stock dataset\footnote{https://www.kaggle.com/szrlee/stock-time-series-20050101-to-20171231?select=GE\_2006-01-01\_to\_2018-01-01.csv} gathers daily stock prices and volume of General Electric stocks. It is a multivariate time series with 5 input features. The final dataset have $3,020$ samples, and a sequence length equal to $40$, corresponding to 2 months of observations (recorded only during business days). 

\subsection{Additional results}
Table~\ref{table:mses-synthetic:sup} presents additional results on the synthetic datasets, and displays the \textit{mse} and \textit{dist-mse} described in section 4.2 of the main paper.  

Table~\ref{table:picp:mpiw:unistep:sup} and Table~\ref{table:picp:mpiw:multistep:sup} present the additional results when doing respectively \textit{unistep forecasting} and \textit{multi-step forecasting}, and display the mean square error over the test set, and the predictive interval metrics (PICP and MPIW) described in section 4.3 of the main paper. For the multivariate time series, the PICP and MPIW are averaged over all the target features. 

\begin{table*}
\caption{Mean Square Error of the mean predictions (mse) and Mean Square Error of the predictive distribution (dist-mse) on the test set versus the ground truth, for Model I and II.
Values are computed with a 5-fold cross-validation procedure on each dataset. 
Std values are displayed in parenthesis when stds $\geq 0.01.$.
For the LSTM and Transformer models with \mcdrop, $p$ is the dropout rate.
For the Bayesian LSTM, $M$ is the number of Monte Carlo samples to estimate the ELBO loss~\cite{fortunato2017bayesian}. We display the results for $M=3$, as it is the default parameter value provided by the blitz library. 
For the \smc, $M$ is the number of particles of the SMC algorithm. 
}
\label{table:mses-synthetic:sup}
\scriptsize
 \centering
\begin{tabular}{c|c|c|c|c}
\toprule
    \multirow{2}{*}{Model} &
      \multicolumn{2}{c}{\textbf{Model I}} &
      \multicolumn{2}{c}{\textbf{Model II}} \\
      & {mse} & {dist-mse} & {mse} & {dist-mse}  \\
      \midrule
     \textbf{True Model}  & 0.5 & 0.50(0.03) & 0.3 & 0.35(0.07) \\
           \midrule
      \textbf{LSTM} & \textbf{0.50} & N.A  & 0.32 & N.A \\
      \textbf{Transformer} & 0.52 & N.A  & 0.32 & N.A \\
      \midrule
      \textbf{LSTM drop.}  &  &   \\
$p = 0.1$    & 0.48 & 0.004  & 0.32 & 0.003 \\
$p = 0.2$    & 0.53 & 0.0099 & 0.34 & 0.007 \\
$p = 0.5$    & 0.53 & 0.03  & 0.33 & 0.02 \\
\midrule
\textbf{Transf. drop.}  &  &  \\
$p = 0.1$    & \textbf{0.50} & 0.02  & 0.31 & 0.03 \\
$p = 0.2$    & 0.50 & 0.02(0.01)  & 0.32 & 0.02 \\
$p = 0.5$    & 0.52(0.01) & 0.05(0.02)  & 0.33(0.02) & 0.05(0.02) \\
\midrule
\textbf{Bayes. LSTM } &  &  \\
$M=3$  &  0.55(0.01) & 0.04 & 0.36(0.01) & 0.05 \\
$M=10$  &  0.53(0.01) & 0.03 & 0.37(0.01) & 0.04 \\
\midrule
\textbf{SMC Transf.}  &  &  \\
$M=10$  &  0.52 & \textbf{0.49} & \textbf{0.30} & \textbf{0.35}
\\
$M=30$  &  0.49 & 0.52 & 0.34 & \textbf{0.35}
\\
\bottomrule
  \end{tabular}
\end{table*}

\begin{table*}
 \centering\hspace*{-5.cm}
\scriptsize
\caption{mse (test loss), PICP and MPIW for unistep forecast. The values in bold correspond to the best performances for each metric.} \label{table:picp:mpiw:unistep:sup}
\label{table:picp:mpiw:unistep}
 \centering\hspace*{-0.6cm}
\resizebox{\textwidth}{!}{
\begin{tabular}{c|c|c|c|c|c|c|c|c|c|c|c} 
\toprule
     & \multicolumn{2}{c|}{{\small Covid}} & \multicolumn{2}{c|}{{\small Air quality}} & \multicolumn{2}{c|}{{\small Weather}} & \multicolumn{2}{c|}{{\small Energy}} & \multicolumn{2}{c}{{\small Stock}} \\
    \hline
  &  mse & picp | mpiw &  mse & picp | mpiw &  mse &  picp | mpiw &  mse & picp | mpiw & mse & picp | mpiw\\
\midrule
\scriptsize
{\small LSTM } & 0.117 & - | - & 0.120 & - | - & 0.076 & - | - & 0.039 & - | - & 0.055 & - | -\\
\midrule
{\small LSTM drop. } &  & & & & & & & \\
$p = 0.1$ & 0.150 & 0.77 | 0.27 & \textbf{0.139} & 0.64 | 0.40 & 0.089 & 0.55 | 0.38 & 0.07 & 0.99 | 0.49 & \textbf{0.065} & 0.93 | 0.28 \\
$p = 0.2$ & 0.159 & 0.79 | 0.38 & 0.152 & 0.72 | 0.54 & 0.101 & 0.64 | 0.53 & 0.103 & 0.99| 0.45 & 0.074 & \textbf{0.95} | \textbf{0.38} \\
$p = 0.5$ & 0.155 & 0.89 | 0.63  & 0.211 & 0.86 | 0.87 & 0.157 & 0.78 | 0.92 & 0.218 & 0.96 | 1.24  & 0.112 & 0.97 | 0.108 \\
\midrule
{\small Transf.} & 0.116 & - | - & 0.132 & - | - & 0.099 & - | - & 0.042 & - | - & 0.068 & - | -\\
\midrule
{\small Transf drop.} & & & & & & & & \\
$p = 0.1$ & \textbf{0.121} & \textbf{0.96} | \textbf{0.37}  & 0.141 & 0.77 | 0.47 & 0.127 & 0.60 | 0.35 & 0.046 & \textbf{0.96} | \textbf{0.45} & 0.076 & 0.93 | 0.34 \\
$p = 0.2$ & 0.129 & 0.94 | 0.46  & 0.159 & 0.89 | 0.58 & 0.133 & 0.72 | 0.44 & 0.053 & 0.98 | 0.64 & 0.082 & 0.95 | 0.41 \\
$p = 0.5$ & 0.208 & 0.97 | 0.76  & 0.196 & \textbf{0.96} | \textbf{0.85} & 0.180 & 0.89 | 0.68 & 0.09 & 0.97 | 1.00 & 0.106 & 0.95 | 0.63 \\
\midrule
{\small Bayesian LSTM}& 0.144 & 0.25 | 0.12 & 0.192 & 0.77 | 0.51 & 0.092 & 0.25 | 0.15 & 0.121 & 0.91 | 0.76 & 0.086 & 0.85 | 0.22 \\
\midrule
{\small SMC Transf.} &  0.128 & 0.997 | 0.70  & 0.148 & 0.97 | 1.54 & 0.180 & \textbf{0.99} | \textbf{1.65} & \textbf{0.043} & 0.99 | 0.82 & 0.071 & 0.99 | 1.08 \\

\bottomrule
\end{tabular}
}
\end{table*}

\begin{table*}
 \centering\hspace*{-5.cm}
\scriptsize
\caption{mse (test loss), PICP and MPIW for multistep forecast. The values in bold correspond to the best performances for each metric.} 
\label{table:picp:mpiw:multistep:sup}
 \centering\hspace*{-0.6cm}
\resizebox{\textwidth}{!}{
\begin{tabular}{c|c|c|c|c|c|c|c|c|c|c|c} 
\toprule
     & \multicolumn{2}{c|}{{\small Covid}} & \multicolumn{2}{c|}{{\small Air quality}} & \multicolumn{2}{c|}{{\small Weather}} & \multicolumn{2}{c|}{{\small Energy}} & \multicolumn{2}{c}{{\small Stock}} \\
    \hline
  &  mse & picp | mpiw &  mse & picp | mpiw &  mse &  picp | mpiw &  mse & picp | mpiw & mse & picp | mpiw\\
\midrule
\scriptsize
{\small LSTM drop.} & & & & & & & & \\
$p = 0.1$ & 0.150 & 0.67 | 0.61 & \textbf{0.139} & 0.54 | 0.73 & \textbf{0.089} & 0.62 | 1.15 & 0.07 & \textbf{0.96}\;|\;\textbf{1.12} & \textbf{0.065} & 0.85 | 0.74 \\
$p = 0.2$ & 0.159 & 0.73 | 0.83 & 0.152 & 0.60 | 0.97 & 0.101 & 0.70 | 1.44 & 0.07 & \textbf{0.96}\;|\;\textbf{1.12} & \textbf{0.065} & 0.85 | 0.74 \\
$p = 0.5$ & 0.155 & 0.80 | 1.64  & 0.211 & 0.70 | 1.31 & 0.157 & 0.75 | 1.74 & 0.218 & 0.88 | 1.57 & 0.112 & 0.87 | 1.42 \\
\midrule
{\small Transf. drop.} & & & & & & & & \\
$p = 0.1$ & \textbf{0.121} & 0.74 | 0.74  & 0.141 & 0.69 | 1.44 & 0.127 & 0.38 | 0.59 & 0.046 & 0.89 | 0.60 & 0.076 & 0.71 | 0.54 \\
$p = 0.2$ & 0.129 & 0.73 | 0.77  & 0.152 & 0.60 | 0.97 & 0.101 & 0.70 | 1.44 & 0.053 & 0.93 | 0.77 & 0.082 & 0.71 | 0.54 \\
$p = 0.5$ & 0.208 & 0.84 | 1.52  & 0.196 & 0.77 | 1.97 & 0.180 & 0.59 | 1.04 & 0.09 & 0.94 | 1.21 & 0.106 & 0.83 | 0.77 \\
\midrule
{\small Bayesian LSTM} & 0.144 & 0.15 | 0.23 & 0.192 & 0.49 | 0.72 & 0.092 & 0.27 | 0.38 & 0.121 & 0.93 | 1.08 & 0.086 & 0.34 | 0.45 \\
\midrule
{\small SMC Transf.} &  0.128 & \textbf{0.91}\;|\;\textbf{1.85}  & 0.148 & \textbf{0.97}\;|\;\textbf{3.17} & 0.180 & \textbf{0.92}\;|\;\textbf{2.90} & \textbf{0.043} & 0.97 | 1.33 & 0.071 & \textbf{0.98}\;|\;\textbf{1.80} \\

\bottomrule
\end{tabular}
}
\end{table*}

\clearpage
\newpage

\bibliographystyle{apalike}
\bibliography{bibliographie}
\end{document}